\documentclass[referee,pdflatex,sn-mathphys-num,Numbered]{sn-jnl}
\usepackage{lineno}
\usepackage{graphicx}%
\usepackage{multirow}%
\usepackage{amsmath,amssymb,amsfonts}%
\usepackage{amsthm}%
\usepackage{mathrsfs}%
\usepackage[title]{appendix}%
\usepackage{xcolor}%
\usepackage{helvet}%
\usepackage{textcomp}%
\usepackage{manyfoot}%
\usepackage{booktabs}%
\usepackage{algorithm}%
\usepackage{algorithmicx}%
\usepackage{algpseudocode}%
\usepackage{listings}%
\usepackage{float}
\usepackage[nolists,tablesfirst,nomarkers]{endfloat}
\DeclareDelayedFloatFlavor{algorithm}{figure}
\usepackage{placeins}
\usepackage{caption}
\usepackage{longtable}%
\usepackage{pdflscape}%
\usepackage{bm}%
\usepackage{array}%
\usepackage{tikz}%
\usepackage{pgfplots}%
\pgfplotsset{compat=1.18}%

\usepackage{hyperref} %
\usepackage{xr-hyper} %
\theoremstyle{thmstyleone}%
\theoremstyle{thmstyletwo}%
\theoremstyle{thmstylethree}%
\newcommand{\supptextref}[1]{\hyperref[#1]{Text~S\ref*{#1}}}
\newcommand{\supptableref}[1]{\hyperref[#1]{Table~\ref*{#1}}}
\newcommand{\suppfigref}[1]{\hyperref[#1]{Figure~\ref*{#1}}}

\begin{document}
\title[Article Title]{A Flexible Adaptive Stable Clustering Algorithm for Archive-Scale Online Mass Spectrometry}

\author[1,2]{\fnm{Shao} \sur{Shi}}
\author*[1,2]{\fnm{Xin} \sur{Yang}}\email{yangx@sustech.edu.cn}
\author[1,2]{\fnm{Huiran} \sur{Feng}}
\author[1,2]{\fnm{Jianhuai} \sur{Ye}}
\author[1,2]{\fnm{Tianlong} \sur{Hu}}
\author[1,2]{\fnm{Yaling} \sur{Zeng}}
\author[1,2]{\fnm{Tzung-May} \sur{Fu}}
\author[1,2]{\fnm{Lei} \sur{Zhu}}
\author[1,2]{\fnm{Huizhong} \sur{Shen}}
\author[1,2]{\fnm{Chen} \sur{Wang}}
\author[1,2]{\fnm{Shu} \sur{Tao}}

\affil*[1]{\orgdiv{Shenzhen Key Laboratory of Precision Measurement and Early Warning Technology for Urban Environmental Health Risks, School of Environmental Science and Engineering}, \orgname{Southern University of Science and Technology}, \orgaddress{\street{1088 Xueyuan Avenue}, \city{Shenzhen}, \postcode{518055}, \state{Guangdong}, \country{China}}}

\affil[2]{\orgdiv{Guangdong Provincial Observation and Research Station for Coastal Atmosphere and Climate of the Greater Bay Area}, \orgname{Southern University of Science and Technology}, \orgaddress{\street{1088 Xueyuan Avenue}, \city{Shenzhen}, \postcode{518055}, \state{Guangdong}, \country{China}}}

\abstract{
Modern online mass spectrometry generates multi-terabyte data streams critical for understanding Earth's environmental systems. However, extracting actionable chemical insights from these repositories is impeded by a computational bottleneck: existing clustering methods force a compromise among scalability, metric flexibility, and algorithmic stability. Here, we introduce Flexible Adaptive Stable Clustering (FASC), a dynamical systems framework that resolves these constraints by architecturally decoupling the similarity kernel from rigorous optimization logic. Unlike legacy heuristics that suffer from stochastic drift and algorithmic blending, FASC employs a Density-Augmented Similarity Selection rule and geometric constraints to guarantee deterministic, order-independent convergence. After validating FASC on canonical machine-learning ground truths (achieving $>99.5\%$ cluster purity and 0.99 Adjusted Rand Index), we deployed the framework on 25 million mass spectra of atmospheric aerosols. Demonstrating strictly linear empirical runtime scaling ($\mathcal{O}(N)$), FASC autonomously mapped atmospheric aging pathways of secondary inorganic aerosols while isolating ultra-rare industrial tracers ($<0.2\%$ abundance), providing a scalable infrastructure for mining environmental big data.

}


\maketitle

\section{Introduction}\label{sec1}
Earth’s atmosphere is a highly dynamic, chemically complex system governed by continuous emissions, multi-phase reactions, and transport. To understand these processes—which dictate climate forcing, cloud formation, and human health—modern environmental monitoring increasingly relies on online mass spectrometry\cite{hollender_nontarget_2017, krauss_lchigh_2010}. Instruments such as online aerosol mass spectrometers (AMS), chemical ionization mass spectrometers (CIMS), and single-particle analyzers now sample ambient environments at sub-second frequencies, generating multi-terabyte data streams over the course of a single field campaign\cite{marx_big_2013}. However, extracting actionable scientific insights from these massive, uncurated data streams is fundamentally bottlenecked by computational limitations.

Unlike offline analytical chemistry, where discrete samples can be matched against curated reference libraries, online mass spectrometry generates millions of unlabeled, highly mixed spectra continuously\cite{schymanski_non-target_2015}. Analyzing this data requires robust, unsupervised clustering engines capable of resolving the continuous chemical evolution of dominant species while simultaneously isolating ultra-rare, transient emission plumes. Yet, these vast spectral repositories remain underexploited because prevailing clustering pipelines buckle at archive scale, forcing researchers to either down-sample their data or rely on aggressive averaging, which inevitably obscures trace chemical signatures and dynamic mixing states\cite{xu_comprehensive_2015, wilkinson_fair_2016}.

Consequently, the field faces an algorithmic \textit{quadrilemma}: current tools force a compromise among \textit{scalability} (execution on massive data streams), \textit{flexibility} (support for diverse, domain-specific chemical metrics), \textit{adaptivity} (autonomous discovery of heterogeneous structures), and \textit{stability} (mathematically reproducible convergence). Researchers lean on a broad catalogue of clustering algorithms, yet prevailing designs remain deficient in addressing these dimensions simultaneously. Prototype models (e.g., K-means, Mini-Batch) require practitioners to prescribe cluster numbers in advance, precluding data-driven adjustment to heterogeneous atmospheric structures\cite{lloyd_least_1982, jain_data_1999}. Density and hierarchical methods offer adaptivity but succumb to quadratic runtime and memory scaling, rendering datasets beyond a few hundred thousand spectra intractable\cite{ester_density-based_1996, campello_hierarchical_2015, murtagh_algorithms_2012}.

Furthermore, environmental spectral clustering is plagued by algorithmic instability. Legacy heuristics long favored in atmospheric science, such as adaptive resonance theory (ART-based algorithms), are notoriously sensitive to the presentation order of the data\cite{carpenter_art_1991, weber_comparison_2016}. This stochastic drift produces "algorithmic blending"—a failure mode where distinct chemical families bleed into overlapping acceptance regions, creating computational phantoms that obscure true physical mixing states\cite{frank_comparative_1998}. While recent high-performance engineering efforts in other domains have accelerated clustering using approximate indexing (e.g., Locality-Sensitive Hashing) or hyperdimensional computing\cite{bittremieux_largescale_2025, wang_mscrush_2018, xu_hyperspec_2023}, these accelerations are often tightly coupled to specific metrics (such as sparse peptide fragment rarity or Hamming distance) that fail to capture the continuous, heavily mixed, and dual-polarity manifolds characteristic of online environmental mass spectrometry.

To resolve these structural bottlenecks, we introduce the Flexible Adaptive Stable Clustering (FASC) algorithm. FASC redesigns high-dimensional clustering from first principles, replacing brittle heuristics with a rigorous dynamical systems framework\cite{frechet_elements_1948}. By architecturally decoupling the definition of chemical similarity from the optimization logic, FASC treats the similarity kernel as an interchangeable module, allowing researchers to deploy diverse metrics without destabilizing the solver. Rather than relying on rigid distance thresholds that arbitrarily fracture continuous data, FASC employs a novel density-augmented similarity selection rule. This mechanism mathematically mimics physical reality: it prioritizes the extraction of dense chemical manifolds (such as the dominant continuum of atmospheric aging) while relegating transient noise to the outlier pool, preventing over-segmentation\cite{rodriguez_clustering_2014}.

Crucially, FASC utilizes Lyapunov-guided structural edits to guarantee deterministic, order-independent convergence, effectively eliminating the algorithmic blending that has historically necessitated months of manual data curation\cite{selim_k-means-type_1984, banerjee_clustering_2004}. We validate FASC on an archive-scale ambient dataset comprising 25 million online mass spectra\cite{shi_technical_2024}, demonstrating execution times that are strictly linear with respect to dataset size. Without prior structural assumptions, FASC successfully mapped the continuous atmospheric processing pathways of secondary inorganic aerosols\cite{dallosto_real_2009} while simultaneously isolating ultra-rare, point-source tracers—such as industrial metal plumes—comprising less than 0.1\% of the population\cite{moffet_chemically_2008, reinard_source_2007}. By bridging the gap between high-performance computing and interpretable chemical pattern recognition, FASC provides a stable, highly scalable digital infrastructure for mining the next generation of online mass spectrometric big data.

\section{Methods}\label{sec2}
\subsection{The FASC Algorithm}

\begin{figure*}[t]
	\centering
	\includegraphics[width=1\textwidth]{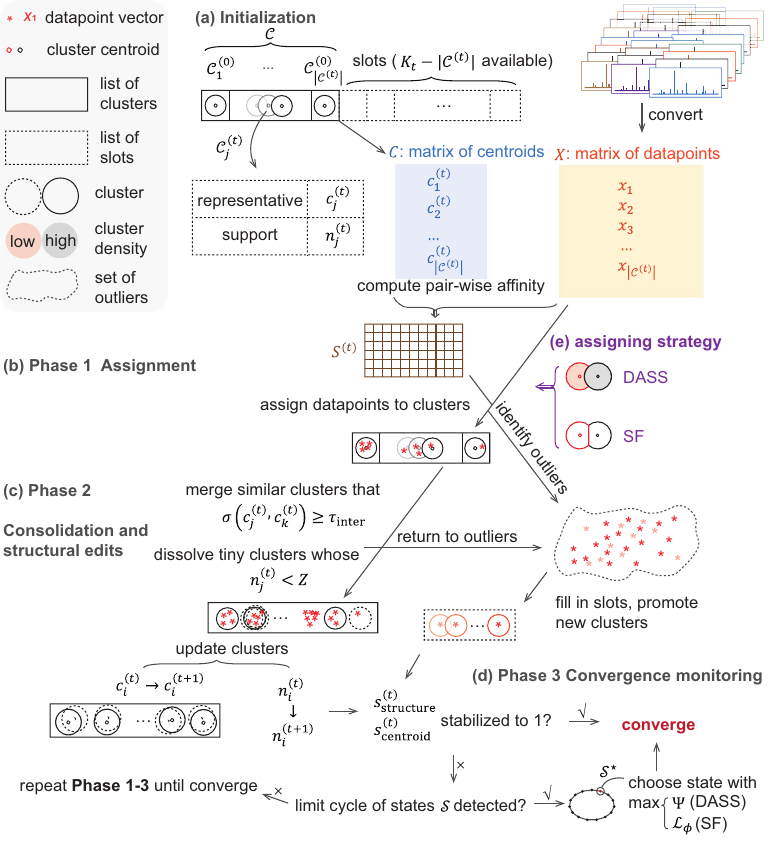}
	\caption{Overview of the FASC workflow. (a) Stratified seeding establishes initial representatives. (b) Streaming assignment scores each spectrum with the dual-cosine kernel. (c) Consolidation updates centroids and performs merge tests. (d) Convergence monitors track cluster size, representative drift, and label changes. (e) Density-augmented similarity selection (DASS) contrasts with the similarity-first (SF) baseline.}\label{fig1}
\end{figure*}

FASC provides the order stability, adaptive sizing, and metric flexibility required for high-dimensional big data (Algorithm~\ref{alg:fasc}, Figure~\ref{fig1}).

\paragraph{Inputs and state.}
Let ${X}=\{x_i\}_{i=1}^N\subset\mathbb{R}^D$ be the preprocessed spectra. The user supplies a similarity kernel $\sigma:\mathbb{R}^D\times\mathbb{R}^D\rightarrow\mathbb{R}$. Four hyperparameters govern the clustering logic: an assignment acceptance threshold $\tau_{\text{intra}}$, a merge threshold $\tau_{\text{inter}}$, a minimum support $Z$, and a capacity schedule $\{K_t\}_{t\geq 0}$ capping active clusters (commonly a single value for all $t$, i.e., $K_t = K_{\max}$, $K_t = N$ for unlimited cluster numbers). Crucially, $K_{\max}$ serves as a budget rather than a rigid structural hypothesis for the maximum number of cluster. The clustering state $\mathcal{S}$ at iteration $t$ is defined as:

\begin{equation}
	\mathcal{S}^{(t)} = \big(\mathcal{C}^{(t)},ID^{(t)}\big), \qquad
	\mathcal{C}^{(t)}=\big\{(c_j^{(t)},n_j^{(t)})\big\}_{j=1}^{|\mathcal{C}^{(t)}|}, \qquad ID^{(t)}\in\{-1,\ldots,|\mathcal{C}^{(t)}|\}^N,
	\label{eq:state}
\end{equation}

where $c_j^{(t)}$ is the representative of the $j^{th}$ cluster, $n_j^{(t)}$ is its support, and $ID^{(t)}$ stores sample-to-cluster assignments ($-1$ denotes provisional outliers). Initialisation sets $\mathcal{C}^{(0)}=\emptyset$ and $ID^{(0)}=-\bm{1}$, with a small seed budget $S_0$ used to instantiate provisional clusters refined during the first pass.

\paragraph{Phase~1: Assignment (Figure~\ref{fig1}b).}
FASC computes affinity scores $S^{(t)}$ of each data vector to each cluster's representative using the similarity kernel $\sigma$. Taking cosine similarity as an example, affinities are calculated via:
\begin{equation}
	S^{(t)} = \widehat{X}\,\widehat{C}^{(t)\top},
	\label{eq:similarity-matrix}
\end{equation}
where $\widehat{X}$ and $\widehat{C}^{(t)}$ are the $\ell_2$-normalised data and representative matrices. The set of admissible clusters for spectrum $x_i$ is:
\begin{equation}
	\mathcal{K}_i^{(t)} = \big\{ j \,\big|\, \sigma(x_i, c_j^{(t)}) \geq \tau_{\text{intra}} \big\}.
	\label{eq:candidate-set}
\end{equation}
If $\mathcal{K}_i^{(t)}=\emptyset$, $x_i$ remains an outlier. Otherwise, FASC assigns the label $w_i^{(t)}$ using one of two rules.

The traditional \textit{similarity-first} (SF) rule, which often fragments natural clusters when data density varies, selects the best match:
\begin{equation}
	\text{SF:}\quad w_i^{(t)} = \arg\max_{j\in\mathcal{K}_i^{(t)}} \sigma(x_i, c_j^{(t)}).
	\label{eq:sf}
\end{equation}
However, the SF strategy guarantees convergence only when the similarity kernel corresponds to a Bregman divergence (e.g., Euclidean distance or Kullback-Leibler divergence), for which the cluster centroid is the unique minimizer of the objective function ~\cite{banerjee_clustering_2004}(\supptextref{supp:text-s5}). For general bounded similarities, SF lacks this theoretical guarantee. Consequently, to overcome this limitation, we introduce the \textit{density-augmented similarity selection} (DASS) rule:
\begin{equation}
	\text{DASS:}\quad w_i^{(t)} = \arg\max_{j\in\mathcal{K}_i^{(t)}} \Big[\sigma(x_i, c_j^{(t)}) + \lambda\,\phi\!\left(n_j^{(t)}\right)\Big],
	\label{eq:dass}
\end{equation}
where $\phi$ is non-decreasing and $\lambda>0$ weights the density contribution (typically $\lambda=1, \phi(n)=n$). Unlike SF, DASS is broadly applicable, requiring only that the similarity kernel be bounded and symmetric. By penalizing the fragmentation of dense clusters, DASS encourages the algorithm to recognize high-density regions as coherent units even amidst scoring fluctuations~\cite{rodriguez_clustering_2014} (Figure~\ref{fig1}e).

While $ID_i^{(t)}$ is updated immediately, support counts remain frozen until Phase~2 to maintain parallel consistency. Finally, if the active cluster count is below $K_t$, a subset of outliers is promoted to singleton clusters to enable adaptive growth. Crucially, because the representative set $\mathcal{C}^{(t)}$ remains frozen during assignment, the calculations for each spectrum $x_i$ are mutually independent. This decoupling renders Phase~1 inherently parallelizable, enabling high-throughput processing via multi-threading or distributed computing—a significant advantage over strictly serial incremental approaches.

\paragraph{Phase~2: Consolidation and structural edits (Figure~\ref{fig1}c).}
Representatives are recomputed from the full memberships of each cluster. For cluster $j$ with indices $I_j^{(t)}$, utilizing a cosine kernel, the update is:
\begin{equation}
	c_j^{(t+1)} = \frac{\sum_{i\in I_j^{(t)}} x_i}{\left\|\sum_{i\in I_j^{(t)}} x_i\right\|_2},
	\qquad
	n_j^{(t+1)} = |I_j^{(t)}|.
	\label{eq:centroid-update}
\end{equation}
For other similarities, $c_j^{(t+1)}$ is updated using the Fr\'echet mean\cite{frechet_elements_1948}, which serves as the generalized centroid for arbitrary metric spaces. Specifically, the Fr\'echet mean corresponds to the arithmetic mean for Euclidean distance, the normalized vector sum for cosine similarity, and the geometric median for Manhattan distance.

Post-update, FASC prunes redundant structure by identifying merge candidates:
\begin{equation}
	\mathcal{M}_j^{(t+1)} = \big\{ k \neq j \,\big|\, \sigma(c_j^{(t+1)}, c_k^{(t+1)}) \geq \tau_{\text{inter}} \big\} \cup \{j\}.
	\label{eq:merge-set}
\end{equation}
Clusters in $\mathcal{M}_j^{(t+1)}$ merge into anchor $j$, and the representative is recomputed. Finally, clusters with $n_j^{(t+1)}<Z$ (excluding newly promoted singletons) are dissolved. Identifiers are renumbered consecutively to prepare for the next iteration.

\paragraph{Phase~3: Convergence monitoring (Figure~\ref{fig1}d).}
The convergence of FASC is grounded in dynamical systems theory, formulated as a block-coordinate optimization over a finite state space. The process is primarily driven by minimizing specific energy landscapes. Specifically, the SF rule optimizes a Lyapunov functional $\mathcal{L}_\phi$ based on Bregman divergence, while the DASS strategy employs a bounded potential function $\Psi$:
\begin{align}
	\mathcal{L}_\phi(\mathcal{S}) &= \sum_{j=1}^{K} \sum_{i \in I_j} D_\phi(x_i,c_j), \label{eq:lyapunov-main} \\
	\Psi(\mathcal{S}) &= \sum_{j=1}^{K} \sum_{i \in I_j} \sigma(x_i,c_j) + \sum_{j=1}^{K} \binom{n_j}{2}. \label{eq:potential-main}
\end{align}

However, the trajectory is not strictly monotonic. Structural edits (merging/splitting) are enforced to satisfy physical constraints ($\sigma(c_j, c_k) < \tau_{\text{inter}}$ and $\sigma(x_i, c_j) > \tau_{\text{intra}}$), occasionally overriding energy descent. While this non-monotonicity effectively prevents entrapment in local minima, it may induce periodic oscillations. Thus, the framework guarantees termination at either a stationary point or a bounded limit cycle~\cite{selim_k-means-type_1984} (\supptextref{supp:text-s5}).

To detect convergence or limit cycles efficiently, we monitor the inter-iteration similarity of centroids and cluster structures:
\begin{align}
	s^{(t)}_{\text{centroid}} &= \sigma\!\left(\text{sort}(n^{(t-1)}), \text{sort}(n^{(t)})\right), \label{eq:size-gap}\\
	s^{(t)}_{\text{structure}} &= \sigma\!\left(C^{(t-1)}, C^{(t)}\right). \label{eq:rep-gap}
\end{align}
This combined metric acts as a sensitive proxy for stability, flagging convergence more rapidly than the objective function itself. If a limit cycle is detected, the algorithm extends execution to record the cycle's history, selecting the specific state $\mathcal{S}^\star$ within the period that globally optimizes $\Psi$ or $\mathcal{L}_\phi$. In practice, tolerance thresholds and early stopping mechanisms are implemented to account for floating-point precision.

\subsection{Validation Strategy and Dimensionality Reduction}

Evaluating unsupervised clustering in high-dimensional online mass spectrometry presents a well-known paradox. Standard internal validation indices (e.g., Euclidean Silhouette Score, Davies-Bouldin) suffer from the "curse of dimensionality," losing their discriminative power regarding chemical structure as spatial distances homogenize\cite{dexter_testing_2016, handl_computational_2005, alexandrov_spatial_2020}. Instead of relying on these degraded \textit{post hoc} metrics, FASC bypasses this limitation by embedding geometric validation directly into its operational logic. The algorithm's dual-threshold architecture—comprising the assignment threshold ($\tau_{\text{intra}}$) and merge threshold ($\tau_{\text{inter}}$)—functions as a continuous, \textit{prescriptive} internal validation mechanism. By mathematically enforcing minimum angular compactness ($\tau_{\text{intra}}$) and strict inter-cluster separation ($\tau_{\text{inter}}$) during runtime, FASC inherently guarantees the structural integrity that standard internal indices attempt to measure.

Furthermore, we refrain from applying external validity indices directly to ambient environmental datasets because ``ground truth" labels in such public repositories are inevitably derived from legacy algorithms or subjective human gating. These traditional annotations systematically conflate distinct mixing states, meaning consistency with external labels does not necessarily imply algorithmic correctness in unsupervised discovery.

To resolve this paradox, we implemented a multi-tiered validation strategy that establishes foundational mathematical accuracy prior to interpreting the unlabeled ambient data:

\textbf{1. Quantitative Ground-Truth Benchmarking:} To provide an objective measure of algorithmic accuracy free from chemical labeling bias, we first benchmarked FASC on the MNIST handwritten digit database ($N=70{,}000$ images, $D=784$ pixel features). MNIST is universally recognized as a foundational baseline dataset for evaluating and comparing machine learning algorithms. We explicitly selected it not only for its established role as the ``gold standard" computational benchmark, but because it provides an absolute, indisputable ground truth (digits 0--9) while presenting structural challenges mathematically analogous to ambient mass spectrometry. Specifically, MNIST data occupies a high-dimensional feature space and forms continuous, non-Euclidean density manifolds characterized by massive intra-class variance (e.g., diverse handwriting styles of the same digit). Successfully clustering MNIST requires an algorithm to respect these heterogeneous, overlapping boundaries without artificially fracturing them—precisely the capability required to map continuous atmospheric aging gradients.  

To rigorously account for this structural adaptivity, where FASC autonomously discovers these physically distinct sub-manifolds within a single broad class, we utilized a strict Majority-Vote Mapping protocol. The algorithm was granted a generous capacity budget ($K_{\max} = 1000$), allowing it to freely form density-based sub-clusters. Each resulting sub-cluster was mapped to its majority ground-truth label, while any unassigned spectra (provisional outliers) were mapped to a generic penalty class ($-1$). 

We evaluated the mapped partitions using overall Classification Accuracy (Cluster Purity), the Adjusted Rand Index (ARI), and Normalized Mutual Information (NMI). Cluster Purity provides an intuitive baseline, calculated as the straightforward fraction of total samples correctly assigned to their true class after majority mapping. However, because high purity can sometimes be trivially achieved in clustering by severely over-segmenting the data, we rely on ARI and NMI for rigorous statistical validation. The ARI evaluates the pairwise agreement between the predicted clusters and ground-truth classes while rigorously correcting for chance groupings. Let $n_{ij}$ denote the number of samples assigned to both true class $i$ and predicted cluster $j$, with $a_i$ and $b_j$ representing the respective row and column marginal sums. The ARI is formally defined as:
\begin{equation}
	\text{ARI} = \frac{\sum_{i,j} \binom{n_{ij}}{2} - \left[ \sum_i \binom{a_i}{2} \sum_j \binom{b_j}{2} \right] / \binom{N}{2}}{\frac{1}{2} \left[ \sum_i \binom{a_i}{2} + \sum_j \binom{b_j}{2} \right] - \left[ \sum_i \binom{a_i}{2} \sum_j \binom{b_j}{2} \right] / \binom{N}{2}},
\end{equation}
where $N$ is the total number of samples. An ARI of 1.0 indicates identical partitions, whereas 0.0 indicates completely random assignments. 

Complementarily, NMI quantifies the statistical dependence between the true class distribution $Y$ and the predicted cluster distribution $C$. It is defined using information theory as:
\begin{equation}
	\text{NMI}(Y, C) = \frac{2 \cdot I(Y; C)}{H(Y) + H(C)},
\end{equation}
where $I(Y; C)$ represents the mutual information between the two distributions, and $H(\cdot)$ denotes the Shannon entropy. The NMI is bounded between 0 (strictly independent) and 1.0 (perfectly correlated), providing a robust measure of cluster purity that is mathematically resilient to structural over-segmentation.

\textbf{2. Topological Validation of Ambient Data:} To evaluate performance on the unlabeled 25-million spectra SPMS dataset, we assessed the physical continuity of the extracted clusters using low-dimensional embeddings generated via t-Distributed Stochastic Neighbor Embedding (t-SNE) and Uniform Manifold Approximation and Projection (UMAP)\cite{becht_dimensionality_2019, linderman_fast_2019}. Linear methods such as PCA (\suppfigref{fig:PCA}) were excluded in our discussion as they are confounded by high-intensity peaks and fail to capture the complex, non-linear manifold structure\cite{hotelling_analysis_1933}. Crucially, the embeddings were computed directly from the raw spectral features entirely independent of the clustering process; FASC assignments were overlaid \textit{post hoc}. We evaluated the stability of these embeddings across a extensive range of hyperparameters (perplexity 15--3000 for t-SNE; nearest neighbors 3--6000 for UMAP)\cite{kobak_art_2019, diaz-papkovich_umap_2019} to confirm that the mathematically isolated clusters reflect robust, continuous chemical manifolds rather than local projection artifacts.

\subsection{The Mass Spectra Dataset and Details of Benchmark}

The FASC algorithm was implemented in MATLAB to maximize matrix-multiplication throughput and optimize memory allocation during the streaming processing of archive-scale datasets. To ensure broad accessibility and eliminate the requirement for a commercial license, a compiled, standalone executable of FASC is freely available at [GitHub Link]. This executable can be run using the free MATLAB Runtime environment and accepts standard tabular data formats, allowing seamless integration into Python or R-based data-processing pipelines. Furthermore, the complete architectural logic and pseudocode are provided in the \supptextref{supp:text-spseudocode} to facilitate future native implementations in open-source ecosystems.

We ran quick benchmark of FASC on a data set involving 25 million mass spectra. Mass spectra were acquired with a single-particle mass spectrometer (SPMS) that rapidly delivers millions of medium-resolution spectra. Instrumental details are provided elsewhere\cite{shi_technical_2024}. Briefly, particles are focused by an aerodynamic lens, their vacuum aerodynamic diameter is derived from the time-of-flight (TOF) between two continuous 532~nm Nd:YAG lasers, and a triggered 266~nm Nd:YAG laser ionises each particle before bipolar TOF detection yields simultaneous positive and negative spectra.

Sampling took place at the Southern University of Science and Technology (SUSTech), Shenzhen, China between 2 and 30 April 2021, generating 24{,}742{,}408 TOF spectra of 12{,}371{,}204 atmospheric particles. To convert the raw time-of-flight signals into high-dimensional vectors suitable for clustering, peaks were automatically extracted, centroided, and calibrated with a standard-free algorithm across the dual-polarity mass range. Each particle's positive and negative spectra were concatenated and binned into discrete mass-to-charge ($m/z$) channels at unit resolution, resulting in a sparse $D$-dimensional feature vector $x$ for each particle. Peak intensities were preserved to reflect relative ion abundances. We ran the benchmark of FASC under varying thresholds ($\tau_{\text{intra}}=\tau_{\text{inter}}=0.7, 0.8, 0.9$, and $0.95$) with a capacity budget of $K_{\max}=50$.

The detailed runtime performance and results is documented in \supptextref{supp:text-s6}. All preprocessing and clustering were performed on the Taiyi computing cluster at SUSTech, using 5 × 1st-generation Intel Xeon Scalable processors (2.1 GHz, 24 cores per CPU, $100$ cores used). Averagely, it takes only 2.3 hours for a whole clustering process on this 25-million data under the current settings (\supptextref{supp:text-s6}). The independence of sample assignments permits extensive parallelisation to reduce wall-clock time of FASC clustering. However, to strictly isolate algorithmic efficiency from hardware acceleration and ensure a fair comparison with single-threaded baselines, our reported experiments restricted the dominant similarity evaluations to a serial workflow, enabling multi-threading only for the auxiliary DASS density computations.

\subsection{A Customized Similarity Algorithm for Online SPMS}
Simultaneous acquisition of positive- and negative-ion spectra in SPMS demands a similarity function that treats both polarity channels cohesively. For any spectrum pair $(u,v)$ we evaluate the conventional cosine similarities $\text{Cosine}_{\text{Pos}}(u,v) = \widehat{u}_{\text{Pos}}^\top\widehat{v}_{\text{Pos}}$ and $\text{Cosine}_{\text{Neg}}(u,v) = \widehat{u}_{\text{Neg}}^\top\widehat{v}_{\text{Neg}}$ on the polarity-wise $\ell_2$-normalised subvectors (degenerate channels contribute a cosine score of zero). The dual-cosine similarity is then defined as
\begin{equation}
	\sigma_{\text{dual cosine}}(u,v) = \min\big(\text{Cosine}_{\text{Pos}}(u,v),\,\text{Cosine}_{\text{Neg}}(u,v)\big).\label{eq:dual-cosine-sim}
\end{equation}
The polarity-aware minimum prevents a strong match in one channel from masking disagreement in the other, so cluster assignments are governed by the least concordant polarity.

The corresponding dual-cosine dissimilarity is the polarity-aware geometry through the angular map
\begin{equation}
	d_{\text{dual cosine}}(u,v) = \arccos\big(\sigma_{\text{dual cosine}}(u,v)\big),\label{eq:dual-cosine-dist}
\end{equation}
which is equivalent to taking the maximum of the polarity-specific angles. Consequently, the penalty reflects the least aligned polarity while remaining bounded within the familiar $[0,\pi]$ interval of cosine geometry. Mathematically, the dual-cosine dissimilarity inherits symmetry and non-negativity from its cosine components, but it violates the identity of indiscernibles ($d_{\text{dual cosine}}(u,v)=0$ can occur even when $u \neq v$). Consequently the construction lies outside both the similarity-metric and divergence families, yet still fits the bounded-kernel, admissible-centroid assumptions underpinning FASC’s convergence guarantees (\supptextref{supp:text-s5}), offering an example that demonstrates flexibility. Because the minimum enforces simultaneous agreement across both polarities, the dual-cosine rule functions as an explicit data-fusion step for the paired spectra recorded per particle. The dissimilarity is also used in visualising the clustering results that are discussed in the subsequent section.

\section{Results}\label{sec3}
\subsection{Scalability and Computational Efficiency}

To assess the suitability of FASC for high-throughput mass spectrometry, we analysed its asymptotic complexity and memory footprint relative to established baselines (detailed derivations in \supptextref{supp:text-complexityFASC} and \supptextref{supp:text-s3}). Let $N$ denote the total number of acquired spectra and $D$ the spectral dimensionality. Theoretical analysis establishes that the total runtime of FASC scales as:
\begin{equation}
	T_{\mathrm{FASC}} = \mathcal{O}\!\left(T_{\mathrm{conv}} \cdot (N K_{\max} + K_{\max}^2) \cdot C_\sigma \right),
\end{equation}
where $K_{\max}$ is the maximum active cluster count and $T_{\mathrm{conv}}$ is the convergence horizon. Crucially, as derived in \supptextref{supp:text-complexityFASC}, the computational cost of structural edits (Phase 2) is strictly bounded by $\mathcal{O}(N D)$ per iteration. This is because the cumulative support of all merged clusters cannot exceed the total dataset size $N$, structurally preventing the combinatorial explosion typical of hierarchical agglomeration. Consequently, since $N \gg K_{\max}$, the quadratic cluster-interaction term vanishes, and the workload is dominated by the linear assignment term $\mathcal{O}(N K_{\max})$. Note that any similarity-based clustering pipeline must evaluate at least one kernel per spectrum, implying a runtime lower bound of $\Omega\!\left(N C_\sigma\right)$; FASC approaches these limits up to constant $K_{\max}$ factors.

\begin{figure*}[t]
	\centering
	\includegraphics[width=\textwidth]{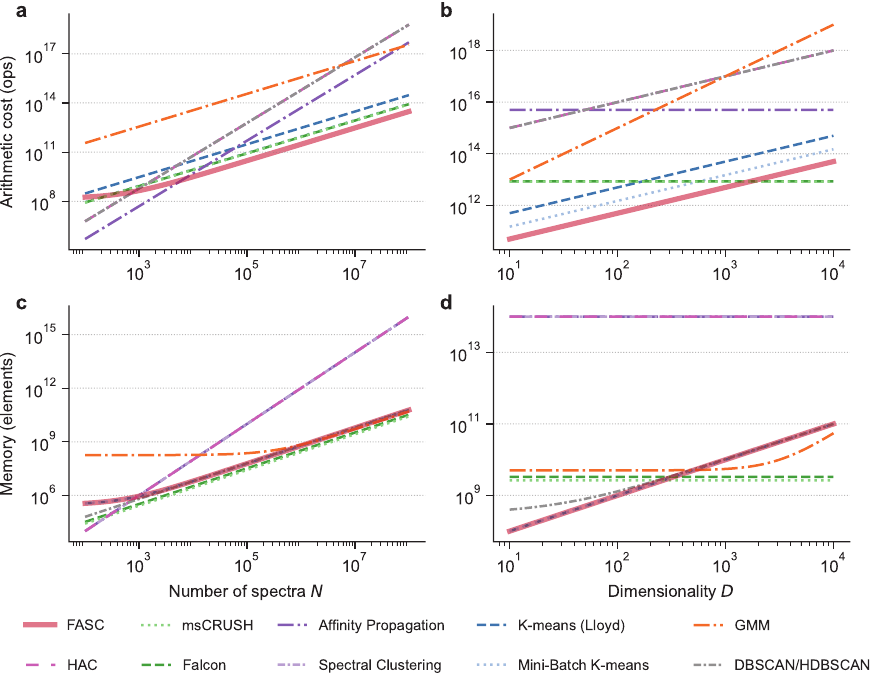}
	\caption{Theoretical scaling envelopes for FASC relative to representative clustering algorithms. (a) Predicted runtime order as a function of spectra count $N$ for fixed dimensionality $D=600$. (b) Predicted memory footprint versus $N$. (c) Predicted runtime as a function of dimensionality $D$ at $N=10^{7}$. (d) Predicted memory footprint versus $D$. Benchmark parameterisations: $K=500$; Mini-Batch K-means (3 passes); ISODATA ($T_{\mathrm{ISO}}=10$); PROCLUS ($s=\lceil 10K \log K \rceil$); BIRCH ($L=256$); Falcon ($L=128, B=32$); msCRUSH ($L=64, B=32, R=2$); MaRaCluster ($C=500$). Tandem spectra sparsified to $P=200$.}
	\label{fig:complexity-panels}
\end{figure*}

Figure~\ref{fig:complexity-panels} delineates the theoretical scaling envelopes of FASC relative to established computational paradigms, revealing three critical performance constraints in current methodologies. First, the quadratic bottleneck inherent in matrix-based approaches—such as Affinity Propagation~\cite{frey_clustering_2007}, Spectral Clustering~\cite{ng_spectral_2001}, and Hierarchical Agglomerative Clustering~\cite{murtagh_algorithms_2012}—mandates the construction of dense $N \times N$ similarity structures. This $\Omega(N^2)$ complexity effectively precludes their application to datasets exceeding $10^5$ spectra. Second, high-dimensional degeneration presents a challenge for density-based algorithms like DBSCAN~\cite{ester_density-based_1996} and HDBSCAN~\cite{campello_density-based_2013}. While theoretically achieving $\mathcal{O}(N \log N)$ efficiency through spatial indexing, \supptextref{supp:text-s3} details how the "curse of dimensionality" in spectral space ($D \gg 100$) erodes spatial index selectivity. This forces neighborhood queries to scan near-linear fractions of the dataset ($|\mathcal{N}_{\epsilon}(x)| \to N$), reverting the practical runtime towards $\mathcal{O}(N^2 D)$~\cite{weber_quantitative_1998}. Third, while specialized mass spectrometry tools such as Falcon~\cite{bittremieux_largescale_2025} and msCRUSH~\cite{wang_mscrush_2018} employ Locality-Sensitive Hashing (LSH) to achieve sub-linear operation counts, these approximate methods necessitate a fundamental trade-off between speed and accuracy. They rely on probabilistic collision guarantees that risk overlooking rare chemical species and require extensive parameter tuning. FASC resolves these limitations by occupying a distinct operational niche: it matches the efficient linear scalability characteristic of K-means while performing exact, rather than approximate, similarity evaluations against the active model.

Regarding implementation, processing $N > 10^7$ spectra often exceeds standard memory capacities. FASC addresses this via a batched streaming architecture. Because Phase~1 assignments are mutually independent, FASC processes spectra in memory-efficient batches, keeping the working footprint at $S_{\mathrm{FASC}} = \mathcal{O}(B \cdot D) + \mathcal{O}(K_{\max} D)$, where $B$ is the batch size. In our production run on 25 million spectra, system telemetry (\supptextref{supp:text-s6}, \supptableref{tab:si_resource_usage}) confirms this linear memory footprint, with peak usage capped at $\approx 247$ GB and negligible swap usage, enabling archive-scale processing on standard high-memory nodes. Furthermore, the validation run completed in just 1.3 hours (\supptableref{tab:si_runtime_by_tau}), demonstrating a throughput of $>5,000$ spectra per second. This design ensures that FASC maintains linear scalability in both runtime and memory (Figure~\ref{fig:complexity-panels}b,d), effectively bridging the gap between the speed of heuristics and the rigor of optimization-based clustering.

To empirically validate these theoretical bounds, we measured the execution time of FASC across increasingly large subsets of the ambient SPMS dataset ($N = 2 \times 10^5$ to $25 \times 10^6$ spectra). Because FASC is a dynamical system where the number of convergence iterations ($T_{\text{conv}}$) naturally fluctuates based on the topological complexity of the sampled subset (e.g., resolving specific boundary overlaps), we evaluated the algorithm's scaling based on the execution time \textit{per iteration}. The empirical time per iteration exhibits a strictly linear relationship with the dataset size ($R^2 > 0.99$, \suppfigref{fig:fasc_scalability}), directly confirming the $\mathcal{O}(N K_{\max})$ theoretical assignment complexity. Macroscopically, the total wall-clock time maintained this near-linear efficiency; the full 25-million spectra dataset required only 41 iterations to reach a stable fixed point, completing in just 1.5 hours (5,478 seconds). System telemetry confirmed a linear memory footprint with peak usage capped at $\approx 247$ GB and negligible swap usage (\supptableref{tab:si_resource_usage}). This empirical performance confirms that FASC effectively bridges the gap between the speed of approximate heuristics and the rigor of deterministic optimization, yielding a throughput exceeding 4,500 spectra per second without hardware accelerators.
\subsection{Flexibility, Adaptivity, and Stability Analysis}
\label{sec:fas-analysis}

\begin{figure}[t]
	\centering
	\includegraphics[width=4.5in]{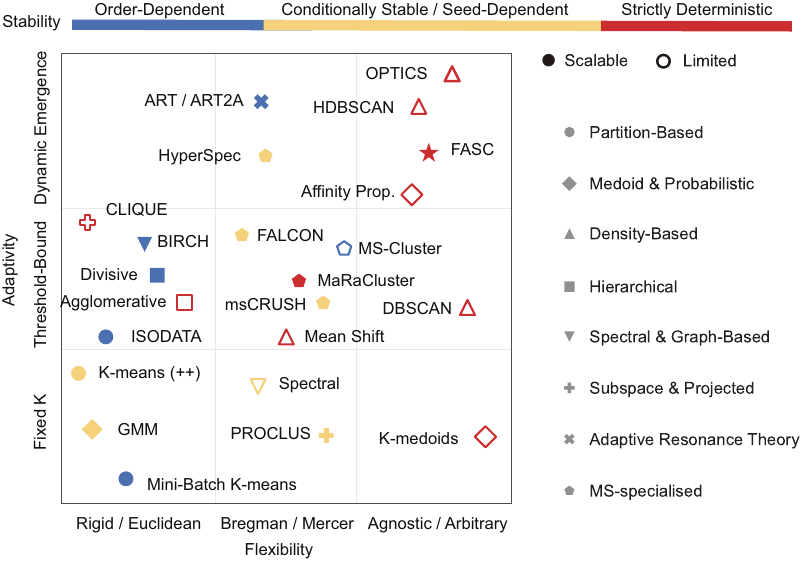}
	\caption{Flexibility--Adaptivity--Stability landscape of mass spectral clustering algorithms. Algorithms are classified according to strict mathematical constraints: \textit{Flexibility} (Euclidean/constrained kernels vs. metric-agnostic oracles), \textit{Adaptivity} (fixed $K$/global thresholds vs. dynamic structural emergence), and \textit{Stability} (order-/seed-dependent variance vs. deterministic permutation invariance). Conventional pipelines (hollow markers) force a structural compromise—prototype models demand rigid Euclidean assumptions, while heuristic density approaches often succumb to threshold limits or stochastic drift (detailed in \supptableref{supp-tab:flex-adapt-ms}). By architecturally decoupling the similarity kernel from a Lyapunov-guided optimization, FASC effectively resolves this compromise, simultaneously achieving metric agnosticism, autonomous structural emergence, and deterministic fixed-point convergence alongside linear scalability.}
	\label{fig:flex-adapt-stability-map}
\end{figure}

To situate FASC within the broader algorithmic landscape and address the historical subjectivity in evaluating clustering performance, we established a rigorous, mathematically objective taxonomy for Flexibility, Adaptivity, and Stability (FAS). As detailed in \supptextref{supp:text-s4} and visualized in Figure~\ref{fig:flex-adapt-stability-map}, we classify canonical clustering pipelines based on strict structural constraints rather than qualitative performance:

\begin{itemize}
	\item \textbf{Flexibility (Kernel Constraint):} Evaluates the mathematical restrictions imposed on the similarity function. Algorithms are classified as \textit{Euclidean} (requiring $\ell_2$-norms, e.g., K-Means), \textit{Constrained} (requiring Bregman divergences or specific hashes, e.g., Spectral, Falcon), or \textit{Agnostic} (accepting arbitrary bounded symmetric matrices).
	\item \textbf{Adaptivity ($K$-Dependence):} Evaluates how the final cluster count ($K_{\text{out}}$) is governed. Algorithms are classified as \textit{Fixed $K$} ($K_{\text{out}} = K_{\text{in}}$), \textit{Threshold-Bound} (governed by rigid global spatial limits, e.g., DBSCAN), or \textit{Dynamic Emergence} (budget-constrained local topology).
	\item \textbf{Stability (Permutation Invariance):} Evaluates mathematical reproducibility under data shuffling $\pi(X)$. Algorithms are classified as \textit{Order-Dependent} ($Alg(\pi_1(X)) \neq Alg(\pi_2(X))$, e.g., ART-2A), \textit{Seed-Dependent} (stochastic initialization), or \textit{Deterministic Fixed-Point} ($Alg(\pi(X)) = Alg(X)$).
\end{itemize}

Under this objective taxonomy, current methodologies face an inherent structural compromise. Prototype-based systems prioritize speed but suffer from rigid Euclidean metric assumptions and fixed $K$ requirements. Density-based approaches improve adaptivity but often succumb to threshold-bound limitations or stochastic instability. FASC distinguishes itself by breaking this structural compromise, simultaneously achieving the highest objective tier across all three dimensions without sacrificing near-linear scalability.

FASC achieves Agnostic Flexibility via a rigorous mathematical decoupling of similarity evaluation from cluster updates. Unlike algorithms bound to specific kernels, FASC treats the similarity function as an interchangeable oracle. This metric agnosticism is grounded in our generalized Fréchet mean update rule\cite{frechet_elements_1948}. Crucially, while traditional optimization rules often strictly require the kernel to be a Bregman divergence to guarantee convergence\cite{banerjee_clustering_2004}, the DASS strategy relaxes this constraint, requiring only that the similarity function be bounded and symmetric. Whether the user selects dual-polarity cosine similarity for atmospheric aerosols or alternative domain-specific scores, the algorithm automatically adapts the centroid computation to the induced geometry of the chosen metric.

FASC achieves Dynamic Adaptivity through its regulated capacity schedule and the DASS assignment logic. In contrast to prototype models that demand a static $K$, or density models that rely on a single global $\varepsilon$ radius, FASC allows the cluster inventory to expand or contract strictly in response to local data distribution. The parameter $K_{\max}$ functions as an upper-bound computational budget, not a mandatory structural hypothesis. When the intrinsic number of clusters $K_{\text{true}} < K_{\max}$, excess slots remain empty, preventing spurious over-segmentation. Conversely, when $K_{\text{true}} > K_{\max}$, the DASS policy integrates an occupancy-dependent term ($\lambda\phi(n_j)$) into the assignment objective. This penalizes the fragmentation of high-density manifolds, allowing FASC to function as a high-pass density filter that prioritizes dominant chemical modes while relegating transient noise to the outlier pool.

Finally, FASC achieves Deterministic Stability by providing a rigorous guarantee of permutation invariance. While high-throughput algorithms (e.g., Mini-Batch K-means, ART-2A\cite{carpenter_art_1991, frank_comparative_1998}) sacrifice reproducibility for speed, FASC enforces order-independence through a block-coordinate optimization framework. By strictly decoupling the assignment phase (Phase 1) from the representative update (Phase 2), the algorithm guarantees that any permutation of the dataset $X$ yields an identical state sequence. Formally, the algorithm ascends a constructed potential function $\Psi$ (under DASS), driving the system toward a geometrically stable configuration where no single sample reassignment can further improve the clustering quality\cite{selim_k-means-type_1984}. In practice, convergence is rapid and robust; the provisional outlier pool contracts by $>80\%$ within the first 16 iterations (\suppfigref{fig:si_outliers}), and the inter-iteration centroid similarity stabilizes to $>0.999$ by the tenth pass (\suppfigref{fig:si_iter_similarity}), confirming that the deterministic assignment protocol rapidly locks onto the global density modes.

\subsection{Quantitative Validation and the Resolution of Algorithmic Blending}

Before applying FASC to the highly mixed, unlabeled topologies of atmospheric mass spectrometry, we validated its foundational mathematical accuracy and metric agnosticism on a canonical high-dimensional dataset with absolute ground-truth labels: the MNIST handwritten digit database ($N=10{,}000$ test images, $D=784$ dimensions). This benchmark serves to prove that FASC's DASS rule correctly isolates continuous data manifolds in high-dimensional space without relying on rigid structural priors. 

To rigorously test the algorithm's adaptivity and resistance to over-segmentation, we provided FASC with an over-capacity budget ($K_{\max} = 1000$), significantly exceeding the true intrinsic number of digit classes ($K_{\text{true}} = 10$). We employed the standard cosine similarity metric to demonstrate the framework's metric agnosticism. Because highly adaptive algorithms naturally identify physically distinct sub-manifolds within a single broad class (e.g., separating distinct handwriting styles of the same digit), we applied a strict majority-vote mapping protocol. 

When configured as a strict high-pass filter ($\tau_{\text{intra}} = \tau_{\text{inter}} = 0.90$), FASC achieved a Classification Accuracy (Cluster Purity) of 99.51\% on the extracted core manifolds, along with an ARI of 0.9975 and a NMI score of 0.9881. 

To provide a rigorous baseline, we evaluated K-means clustering (using the identical cosine metric) across multiple capacity configurations. We first constrained K-means to the true intrinsic class count ($K=10$). Under this rigid hypothesis, K-means achieved a mapping accuracy of 60.37\% (ARI = 0.4426, NMI = 0.5375). This reflects a fundamental mathematical limitation in rigid prototype models: forcing highly diverse, non-Euclidean sub-manifolds into a fixed number of spherical averages structurally obscures intra-class variance. To ensure an equivalent comparison, we subsequently granted K-means expanded budgets of $K=300$ and $K=1000$, applying the identical majority-vote mapping protocol used for FASC. While its performance improved, K-means plateaued at 95.41\% accuracy (ARI = 0.9029, NMI = 0.8902) even under the 1000-cluster budget. 

The performance differential between K-means and FASC, despite operating under identical capacity budgets and metrics, highlights the utility of FASC’s dual-threshold architecture. K-means mandates the assignment of 100\% of data points—including boundary noise and ambiguous transitions—into clusters, inevitably compromising the purity of the core manifolds. In contrast, FASC's thresholds ($\tau_{\text{intra}}$ and $\tau_{\text{inter}}$) function as a tunable density filter. Data points that fail to meet the $\tau_{\text{intra}}$ compactness requirement are autonomously identified as transitional outliers, protecting the pure cluster cores. By modulating these thresholds, the user directly controls the precision-recall trade-off: relaxing the threshold (e.g., $\tau = 0.79$) increases overall data recall but reduces the core purity to 65.01\% (ARI = 0.4668) as distinct classes begin to mathematically overlap in angular space. This tunable behavior reflects the topological phase transition required for environmental mass spectrometry: lower thresholds allow researchers to map continuous atmospheric mixing states, while stringent thresholds guarantee the $>99.5\%$ purity required to isolate rare, pristine point-source tracers from background noise.

These quantitative metrics are directly attributable to FASC's continuous, \textit{prescriptive} internal validation mechanism. Standard internal validation metrics evaluate clustering quality \textit{post hoc} by measuring intra-cluster compactness against inter-cluster separation. FASC, however, embeds these geometric principles directly into its optimization loop. The assignment threshold ($\tau_{\text{intra}}$) mathematically guarantees minimum angular compactness, while the merge threshold ($\tau_{\text{inter}}$) enforces strict inter-cluster separation dynamically during Phase 2 consolidations. 

This prescriptive geometry, combined with a strict decoupling of assignment logic from representative updates, fundamentally resolves the persistent "algorithmic blending" problem inherent in legacy atmospheric clustering tools. While recent high-performance algorithms such as Falcon and MaRaCluster offer rapid execution, they are architecturally optimized for proteomics workflows; their reliance on precursor-mass windowing, peptide fragment-rarity scoring, and specific hash approximations renders them structurally incompatible with the continuous, un-windowed, and dual-polarity nature of online environmental mass spectrometry. Consequently,ART-based algorithms like ART2A remain the canonical baseline for online single-particle MS clustering. 

However, ART2A relies on a single "vigilance" parameter to govern assignment and lacks a retroactive structural constraint. Because centroids update sequentially in online ART models, the final partition exhibits high sensitivity to the data stream's arrival order. More critically, centroids frequently drift into neighboring acceptance radii, causing their hyperspherical boundaries to overlap. This boundary collapse produces algorithmic blending—a classification artifact where distinct chemical families overlap into one another, requiring researchers to manually consolidate redundant clusters\cite{holecek_analysis_2007, frank_comparative_1998}. 

FASC eliminates this artifact through its iterative batch architecture. During Phase 1, assignments for all $N$ samples are computed against a static set of representatives, neutralizing sample-order bias. During Phase 2, FASC physically prohibits overlapping acceptance regions; if any centroids drift beyond the $\tau_{\text{inter}}$ threshold after recomputation, they are rigorously merged. 

\begin{figure}[t]
	\centering
	\includegraphics[width=0.9\textwidth]{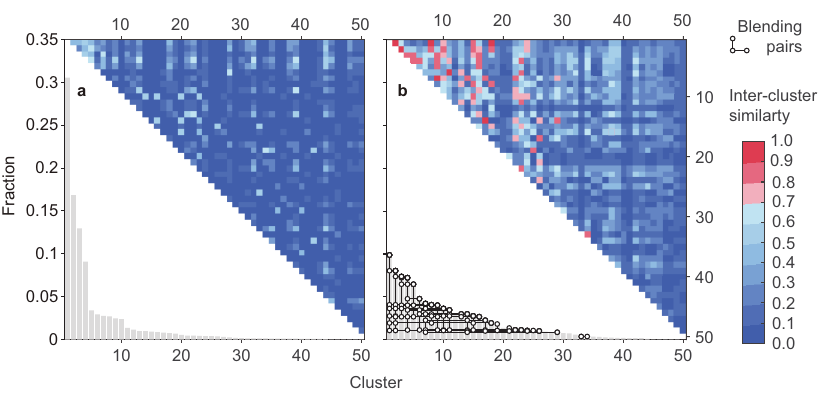}
	\caption{Comparison of algorithmic blending and cluster separation between (a) FASC and (b) ART2A. For a fair comparison, we set the cluster budget $K = 50$, and $\tau = vigilance = 0.8$ for both algorithms. 
		The diagonal bar plots display the population fraction of each cluster relative to the total dataset.
		The upper-triangular heatmaps visualize the pairwise inter-cluster similarity, where blue hues indicate effective separation and red hues indicate high similarity exceeding $\tau_{\text{inter}}$ in FASC or vigilance in ART2A.
		The lower-triangular sections highlight "blending pairs" using scatter points connected by lines, marking instances where centroids fall within neighboring acceptance regions.}
	\label{fig:BlendingTest}
\end{figure}

To empirically demonstrate this theoretical advantage, we evaluated the algorithmic separation of FASC against ART2A using 1.2 million spectra pairs randomly sampled from the ambient SPMS dataset (\suppfigref{fig:BlendingTest}). While ART2A exhibits extensive blending—evidenced by red hotspots and a network of connected pairs indicative of fractured, overlapping vigilance regions—FASC maintains strict mathematical separation with no inter-cluster similarity exceeding $\tau_{\text{inter}}$. It is crucial to distinguish this computational blending from the physical reality of chemical mixing states. While atmospheric particles frequently carry material from multiple sources, FASC's disjoint centroids ensure that such particles are assigned based on their dominant signal or designated as outliers, rather than being obscured by mathematically redundant clusters. By eliminating these computational artifacts, FASC provides a direct, order-robust chemical view of the aerosol population without the \textit{post hoc} merging burden typical of legacy algorithms\cite{zhang_birch_1996, carpenter_art_1991}.

\subsection{Archive-Scale Mining of Atmospheric Mixing States and Rare Tracers}

Applying FASC to the 25 million polarity-resolved single-particle mass spectra (12.37 million particles) yields a comprehensive partition that quantitatively resolves the ambient aerosol population with high  dynamic range. By processing the entire archive-scale dataset simultaneously without resorting to aggressive down-sampling, FASC recovers both the dominant background aerosol and ultra-rare emission sources without manual gating. As detailed in the comprehensive chemical census (\supptextref{supp:text-MSs}, \supptableref{supp-tab:cluster-ms-summary}), FASC reveals that Secondary Inorganic Aerosol (SIA) dominates the population, accounting for 52.7\% of classified spectra (primarily Clusters 1, 19, and 23)\cite{dallosto_chemical_2006, moffet_chemically_2008, reinard_source_2007}. 

\begin{figure}[t]
	\centering
	\includegraphics[width=0.9\textwidth]{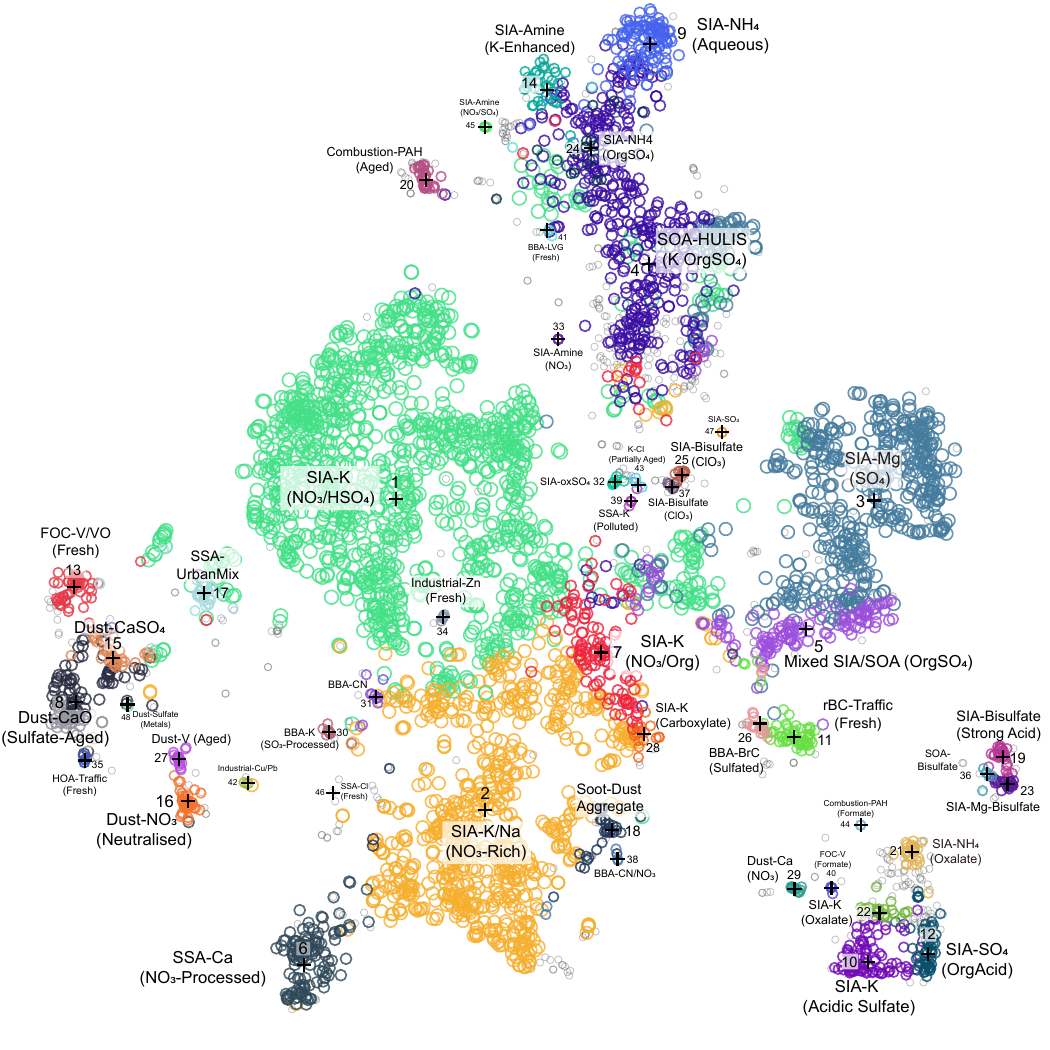}
	\caption{Label-free t-SNE embedding (perplexity 50) of the 12-million-particle ambient atmospheric dataset, overlaid \textit{post hoc} with FASC cluster assignments (dual-cosine metric, $\tau = 0.7$). Each point represents 2{,}000 dual-polarity spectra. Centroids are marked with a $+$. Annotations denote the aerosol category derived from the centroid mass spectra, illustrating FASC's ability to simultaneously map the continuous aging gradient of the SIA backbone while isolating ultra-rare, distinct primary point sources.}
	\label{fig:tsne_perplex50}
\end{figure}

Crucially, because FASC employs the DASS rule and mathematically guarantees inter-cluster separation, it successfully maps the highly complex "SIA backbone" without succumbing to the severe algorithmic blending that typically plagues ART-based analyses of aged aerosols. As visualized in the label-free t-SNE embedding (Figure~\ref{fig:tsne_perplex50}), this backbone represents a continuous atmospheric aging gradient that grades smoothly from nitrate-rich traffic mixtures to highly oxidized, acidic organosulfates\cite{hatch_measurements_2011, bzdek_amine_2010}. While legacy algorithms (such as the ART-based) tend to artificially fracture this continuum into dozens of overlapping, redundant clusters, FASC preserves the intrinsic topological continuity of the secondary aerosol manifold.

Simultaneously, the algorithm demonstrates the "needle in a haystack" capacity to isolate exceptionally rare, distinct primary chemical signals buried within this massive secondary background. FASC successfully extracted distinct topological filaments corresponding to specific minor regimes, including residual oil combustion characterized by distinct V$^{+}$/VO$^{+}$ markers (Cluster 13, 1.25\%) and extremely rare zinc-rich industrial plumes (Cluster 34, $<0.2\%$ abundance)\cite{healy_characterisation_2009, healy_sources_2012}. The ability to autonomously isolate a 0.2\% trace signal from a 12-million-particle background confirms that FASC's deterministic clustering prevents the dilution of rare tracers into the dominant matrix. 

Extensive sensitivity analyses demonstrate that this topological structure—a central continuum decorated by source-specific branches—is a robust physical feature of the atmospheric system rather than a dimensionality-reduction artifact. The "SIA backbone" and its extended filaments persist across a massive range of embedding parameters, remaining stable across t-SNE perplexities from 15 to 3000 (\suppfigref{fig:tSNEs}) and UMAP neighbor counts from 3 to 6000 (\suppfigref{fig:UMAPs}).

Furthermore, FASC offers a tunable computational lens into the hierarchical nature of aerosol mixing states through the similarity thresholds. We observe a distinct topological phase transition: while the moderate threshold utilized here ($\tau = 0.7$, Figure~\ref{fig:tsne_perplex50}) maximizes data recall to map continuous atmospheric processing pathways, strictly increasing $\tau$ ($\ge 0.9$, \suppfigref{fig:tSNEs_0d9}--\ref{fig:tSNEs_0d95}) fractures this continuum into a discrete "archipelago" of high-purity fingerprints. This transition is not a computational failure but a functional feature for exploring internal mixing states. Lower thresholds reveal the aging gradients where chemical signatures blend (e.g., K-rich aerosols coated with secondary sulfates, comprising 22.2\% of spectra\cite{bi_mixing_2011, qin_impact_2006, silva_size_1999}), while higher thresholds act as high-precision extractors for chemically pristine tracers. 

The spectral fingerprints underpinning these structures are thoroughly documented in the centroid atlas (\suppfigref{fig:si_ms_centroids}), confirming that the partition tracks chemically rigorous atmospheric phenomena. For instance, marine-influenced clusters (3.8\%) are correctly identified by combinations of Na$^+$ and K$^+$ with processed chloride loss\cite{gard_direct_1998, holecek_analysis_2007, pratt_observation_2010}, clearly separated from the K-rich urban background. Taken together, these results confirm that the metric-agnostic, dual-threshold architecture of FASC bridges holistic atmospheric processing pathways and targeted trace-element detection, converting massive streams of online sensor data into interpretable atmospheric chemistry.

\section{Discussion}\label{sec4}
FASC demonstrates that the historical trade-offs among \textit{scalability}, \textit{flexibility}, \textit{adaptivity}, and \textit{stability} in high-dimensional clustering are not fundamental limits, but rather artifacts of coupled algorithmic designs\cite{xu_comprehensive_2015, jain_data_1999}. By architecturally decoupling the definition of similarity from the optimization logic, FASC effectively resolves this "algorithmic quadrilemma." The framework establishes a rigorous protocol where the similarity kernel functions as an interchangeable oracle, ensuring the objective flexibility to deploy arbitrary metrics without destabilizing the solver. Whether utilizing the dual-polarity cosine metric for online single-particle mass spectrometry, or alternative domain-specific distance oracles, FASC seamlessly adapts its Fréchet mean update to the induced geometry of the data. This contrasts sharply with legacy heuristics like ART or modern approximate indexing approaches, which frequently achieve processing speed by sacrificing deterministic stability, theoretical robustness, or flexibility to continuous, non-Euclidean manifolds\cite{bittremieux_largescale_2025, carpenter_art_1991}.

A critical innovation in FASC is the DASS rule, which resolves the tension between adaptivity and noise resilience. Unlike traditional algorithms that blindly force all data into clusters or rely on rigid global spatial search radii, DASS incorporates local density information ($\lambda \phi(n_j)$) into the assignment objective. This mechanism mathematically mimics physical reality: it recognizes that meaningful chemical species tend to form dense manifolds, whereas instrumental noise and transitional mixing states are sparse\cite{rodriguez_clustering_2014}. As demonstrated by our quantitative MNIST benchmarking, granting FASC a massive capacity budget ($K_{\max} = 1000$) did not result in spurious over-segmentation. Instead, DASS acted as a high-pass density filter, autonomously extracting the true structural sub-manifolds with 99.51\% purity while relegating ambiguous boundaries to the outlier pool. 

In the unsupervised mining of archive-scale atmospheric data, where absolute "ground truth" labels are non-existent, this structural purity and algorithmic stability become the primary metrics of validity\cite{handl_computational_2005}. Our analysis reveals that legacy algorithms heavily utilized in atmospheric science suffer from extensive "algorithmic blending"—a classification flaw where stochastic, sequentially updated centroids drift into overlapping acceptance regions\cite{frank_comparative_1998}. By enforcing strict inter-cluster separation dynamically via its dual-threshold architecture ($\tau_{\text{intra}}$ and $\tau_{\text{inter}}$), FASC utilizes a prescriptive internal validation mechanism that physically prohibits these overlapping boundaries. Furthermore, its Lyapunov-guided block-coordinate optimization guarantees mathematically deterministic convergence to a fixed point or bounded limit cycle\cite{selim_k-means-type_1984, banerjee_clustering_2004}. This ensures that any chemical mixing state identified by FASC is a genuine physical property of the atmospheric system, not a computational phantom generated by input order or random seeding.

The practical power of this stability is exemplified by FASC's unprecedented resolution of the 25-million spectra dataset. The algorithm autonomously recovered the "SIA backbone"—a continuous chemical manifold spanning from nitrate-rich traffic emissions to highly aged, acidic organosulfates\cite{hatch_measurements_2011, healy_sources_2012}. Crucially, by modulating the $\tau$ thresholds, FASC offers researchers direct control over the precision-recall trade-off governing atmospheric topologies. Lower thresholds successfully mapped the continuous gradients of internal mixing (e.g., K-rich aerosols coated with secondary sulfates), while higher thresholds triggered a topological phase transition, shattering the continuum into an "archipelago" to isolate highly specific, ultra-rare tracers. This enabled the simultaneous extraction of the massive background continuum and trace-level primary signals, such as vanadium-rich fuel residues (1.25\% abundance) and pristine industrial copper/zinc particles ($<0.2\%$ abundance)\cite{moffet_chemically_2008, healy_characterisation_2009}. 

Finally, FASC transitions these rigorous mathematical guarantees to the archive scale. As empirically validated across multi-million spectra subsets, the algorithm operates in strictly linear execution time ($\mathcal{O}(N)$ per iteration) with a linear memory footprint. This exceptional efficiency enabled the clustering of 25 million high-dimensional particles in just 1.5 hours on standard high-memory computing nodes. By providing both HPC-optimized streaming implementations and open-source Python frameworks, FASC seamlessly integrates into modern data-science pipelines. Ultimately, by anchoring metric flexibility and structural adaptivity in deterministic, analyzable operators, FASC resolves the computational bottleneck of online mass spectrometry, converting the deluge of high-throughput sensor data into reproducible, actionable scientific insight.

\bmhead{Supplementary information}
The Supplementary Information is provided as Texts~S1--S10 (Available at https://doi.org/10.5281/zenodo.20080723). \supptextref{supp:text-s1} provides an overview of the supplementary analyses,  \supptextref{supp:text-s2} documents the dimensionality-reduction embeddings for qualitative validation, \supptextref{supp:text-s3} profiles the computational footprint of conventional baselines, and \supptextref{supp:text-s4} evaluates their flexibility, adaptivity, and stability with respect to FASC.
 \supptextref{supp:text-spseudocode} presents the full algorithm pseudocode, \supptextref{supp:text-s5} formalises the convergence guarantees, \supptextref{supp:text-s_scalability} reports the scalability benchmark, \supptextref{supp:text-complexityFASC} derives the computational complexity bounds, \supptextref{supp:text-s6} reports the performance on the 25 million spectrum run, and \supptextref{supp:text-MSs} curates the centroid-level mass-spectral fingerprint analysis.

\bmhead{Acknowledgements}
This work was supported by the National Natural Science Foundation of China (Grant Nos. 42530609 and 41827804), the Ministry of Science and Technology of the People's Republic of China (Grant No. 2023YFE0112900), the Shenzhen Science and Technology Program (Grant No. KCXFZ20230731093601003), the Shenzhen Basic Research Program (Grant No. JCYJ20241202152804007), and the Shenzhen Peacock Team Project (High Spatiotemporal Resolution Monitoring and Health Warning Platform for Urban Atmospheric Environment). We also acknowledge the Center for Computational Science and Engineering at SUSTech for providing the high-performance computational resource.

\section*{Data and Code Availability}

\textbf{Code Availability:} 
The FASC algorithm is open-source and released under the GNU General Public License v3.0 (GPLv3). To ensure maximum accessibility and utility across the scientific community, the framework is provided in two native implementations: a Python version optimized for seamless integration into modern open-science machine learning pipelines, and a MATLAB version designed to maximize matrix-multiplication throughput and memory mapping during the batched streaming of archive-scale data on high-performance computing (HPC) clusters. To completely eliminate the barrier of commercial software licenses, a compiled, standalone executable is also provided. The complete source code (Python and MATLAB), standalone executable, synthetic benchmarking scripts, and interactive Jupyter tutorials are freely available on GitHub (\url{https://github.com/s129136908794904/FASC/}) and persistently archived on Zenodo (DOI: \href{https://doi.org/10.5281/zenodo.17844844}{10.5281/zenodo.17844844}). Furthermore, the complete architectural logic and pseudocode are documented in the Supplementary Information (\supptextref{supp:text-spseudocode}).

\textbf{Data Availability:}
The complete high-dimensional atmospheric dataset (comprising 25 million single-particle mass spectra collected in Shenzhen) used to empirically validate the algorithm at archive scale is publicly hosted and persistently archived on Zenodo (DOI: \href{https://doi.org/10.5281/zenodo.17788367}{10.5281/zenodo.17788367}). A curated subset of this data is provided directly within the GitHub repository as a demonstration dataset for immediate algorithmic testing and reproducibility. The MNIST handwritten digit database used for the quantitative ground-truth structural benchmarking is a canonical public machine learning dataset, widely accessible via standard open-source repositories.

\processdelayedfloats 
\pagebreak
\bibliography{bibliography}

\end{document}